\documentclass[10pt,journal,compsoc]{IEEEtran}
\newif\ifpeerreview

\peerreviewfalse

\usepackage[nocompress]{cite}
\usepackage{hyperref}
\usepackage{url}
\usepackage{graphicx}   
\usepackage{amsmath}
\usepackage{amssymb}
\usepackage{amsthm}
\usepackage{booktabs}
\usepackage{algorithm}
\usepackage{algorithmic}
\usepackage{tikz}
\usepackage{caption}
\usepackage{subcaption}
\usepackage{float}
\usepackage{stfloats}

\usepackage{lipsum} 

\usepackage[switch]{lineno}

\title{A New Perspective To Understanding Multi-resolution Hash Encoding For Neural Fields}

\author{Steven Tin Sui Luo
\IEEEcompsocitemizethanks{\IEEEcompsocthanksitem STS Luo is with the Department
of Computer Science, University of Toronto, Toronto,
On.\protect\\
E-mail: stevents.luo@mail.utoronto.ca}
}

\begin{document}

\IEEEtitleabstractindextext{%
\begin{abstract}
Instant-NGP has been the state-of-the-art architecture of neural fields in recent years. Its incredible signal-fitting capabilities are generally attributed to its multi-resolution hash grid structure and have been used and improved in numerous following works. However, it is unclear how and why such a hash grid structure improves the capabilities of a neural network by such great margins. A lack of principled understanding of the hash grid also implies that the large set of hyperparameters accompanying Instant-NGP could only be tuned empirically without much heuristics. To provide an intuitive explanation of the working principle of the hash grid, we propose a novel perspective, namely \textbf{domain manipulation}. This perspective provides a ground-up explanation of how the feature grid learns the target signal and increases the expressivity of the neural field by artificially creating multiples of pre-existing linear segments. We conducted numerous experiments on carefully constructed 1-dimensional signals to support our claims empirically and aid our illustrations. While our analysis mainly focuses on 1-dimensional signals, we show that the idea is generalizable to higher dimensions. Our code is publicly available \href{https://github.com/stevolopolis/CP}{here}.
\end{abstract}

}

\maketitle

\IEEEraisesectionheading{
  \section{Introduction}\label{sec:introduction}
}
%
%
%
%
\IEEEPARstart{I}{nstant-NPG} (NGP) \cite{muller2022instant} took the neural radiance field (NeRF) community by storm in 2022 with its incredible encoding efficiency and quality. With the addition of multi-resolution hash grids and carefully integrated CUDA hardware acceleration, NGP exceeded previous state-of-the-art performances by a far margin in general signal fitting tasks and reduced the inverse rendering speeds of large scenes from hours to seconds. Following the initial release of NGP, numerous efforts were attempted to further the boundaries of this line of work \cite{takikawa2023compact} \cite{vqad2022} \cite{li2023neuralangelo} \cite{barron2023zip}. While the dominance of neural field methods led by NGP for inverse rendering has recently been dethroned by explicit methods such as 3D Gaussian Splatting \cite{kerbl20233d}, it is believed that neural field-based methods are still inherently better in areas such as surface reconstruction and meta-learning. 

The success of NGP is generally attributed to its multi-resolution hash grid structure. The grid learns position-dependent feature vectors, which relieves most of the learning burden from the MLP, where expressivity and inference speed are usually bottlenecked due to its fully-connected sequential processing nature. The use of a hash table to store the feature vectors also allows NGP to have a limited memory footprint despite learning fairly dense features for each grid resolution. Yet, imposing an additional grid structure on neural fields creates numerous additional hyperparameters. TABLE \ref{tab:ngp_params} presents the parameters for the grid itself, which is required on top of those for the MLP. While the original manuscript \cite{muller2022instant} presented some default values for their tested use cases such as NeRF, SDFs, and megapixel images, no robust heuristics were provided for choosing this large set of parameters. Network parameters are even directly ignored and assumed to be held fixed. While in practice this is mostly sufficient with experimental data, we believe that it is important for the community to develop a more rigorous understanding of NGP from the ground up and analyze how these different parameters interact with each other. This may also unlock some insights into how we could optimally choose these parameters given a different type of signal and even inspire some new improvements to the current hash grid structure.

\begin{figure}[!h]
    \centering
    \includegraphics[width=.9\linewidth]{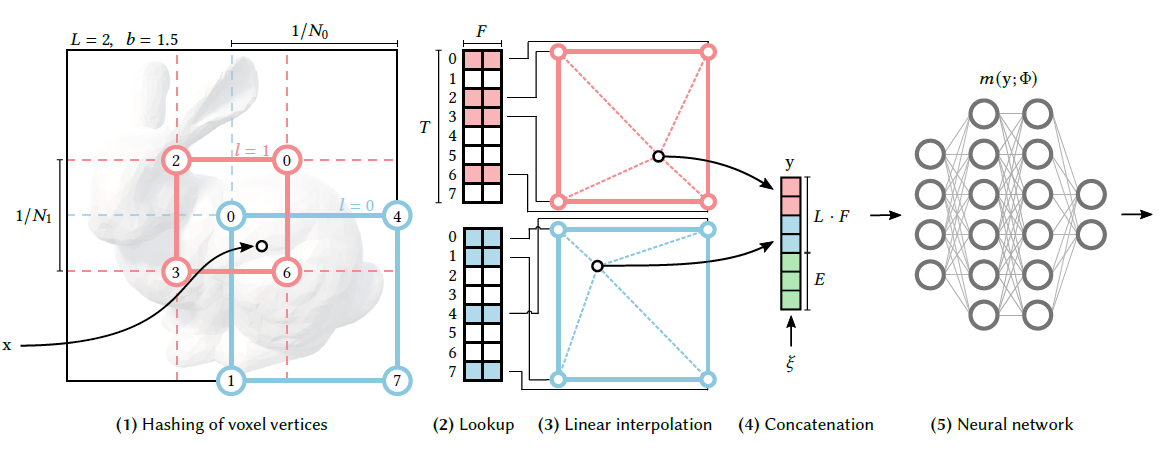}
    \caption{NGP architecture. Taken from \cite{muller2022instant}.}
    \label{fig:ngp_arch}
\end{figure}

\begin{table}[h!]
    \centering
    \begin{tabular}{llr}
        \toprule
        Parameter & Symbol & Value \\
        \midrule
        Number of levels & $L$ & $16$ \\
        Max. entries per level (hash table size) & $T$ & $2^{16}$ to $2^{24}$ \\
        Number of feature dimensions per entry & $F$ & $2$ \\
        Coarsest resolution & $N_{min}$ & $16$ \\
        Finest resolution & $N_{max}$ & $512$ to $524288$ \\
        \bottomrule
    \end{tabular}
    \caption{Hash encoding parameters and default values}
    \label{tab:ngp_params}
\end{table}

It should be noted that a theoretical analysis of a neural network architecture usually involves two aspects: an understanding of the architecture's theoretical \textbf{expressivity}, and an understanding of the architecture's \textbf{training dynamics}. While the former focuses on the relationship between network parameters and their maximum ability to learn signals, the latter focuses on the convergence behaviors of parameters to the global minimum. Notable works for network expressivity include the universal approximation theorem \cite{hornik1989multilayer} and studies that relate a neural network's expressity with its depth/width \cite{ben2023exploring}\cite{raghu2017expressive}\cite{yarotsky2017error}, while that of training dynamics include the developments of neural tangent kernel (NTK) theory \cite{jacot2018neural}. In this work, we focus our analysis under the context of expressivity to allow for a better understanding from the group up.

To analyze the working principles of NGP, we break the model architecture into four components: (1) the grid; (2) hashing; (3) multi-resolution; (4) the MLP. This paper in particular proposes a novel perspective that aids the understanding of how the grid increases the expressivity of the vanilla MLP. We name this perspective ``\textbf{domain manipulation}''. We show with a set of experiments using 1-dimensional toy signals the method, extent, and constraints of how ``\textbf{domain manipulation}'' could improve the network's expressivity. This provides us with a principled way to understand how the grid learns features from the ground up, and explains the necessity of having either higher dimensional features or a multi-resolution structure. We note that numerous subsequent efforts advanced the use of grids without (2) and (3) such as \cite{Fridovich-Keil_2023_CVPR} \cite{cao2023hexplane} \cite{chen2022tensorf} and our insights about the grid could also be extended to aid the understanding of these methods.


\begin{figure*}[!htp]
   \centering
    \begin{subfigure}{.475\linewidth}
        \includegraphics[width=\linewidth]{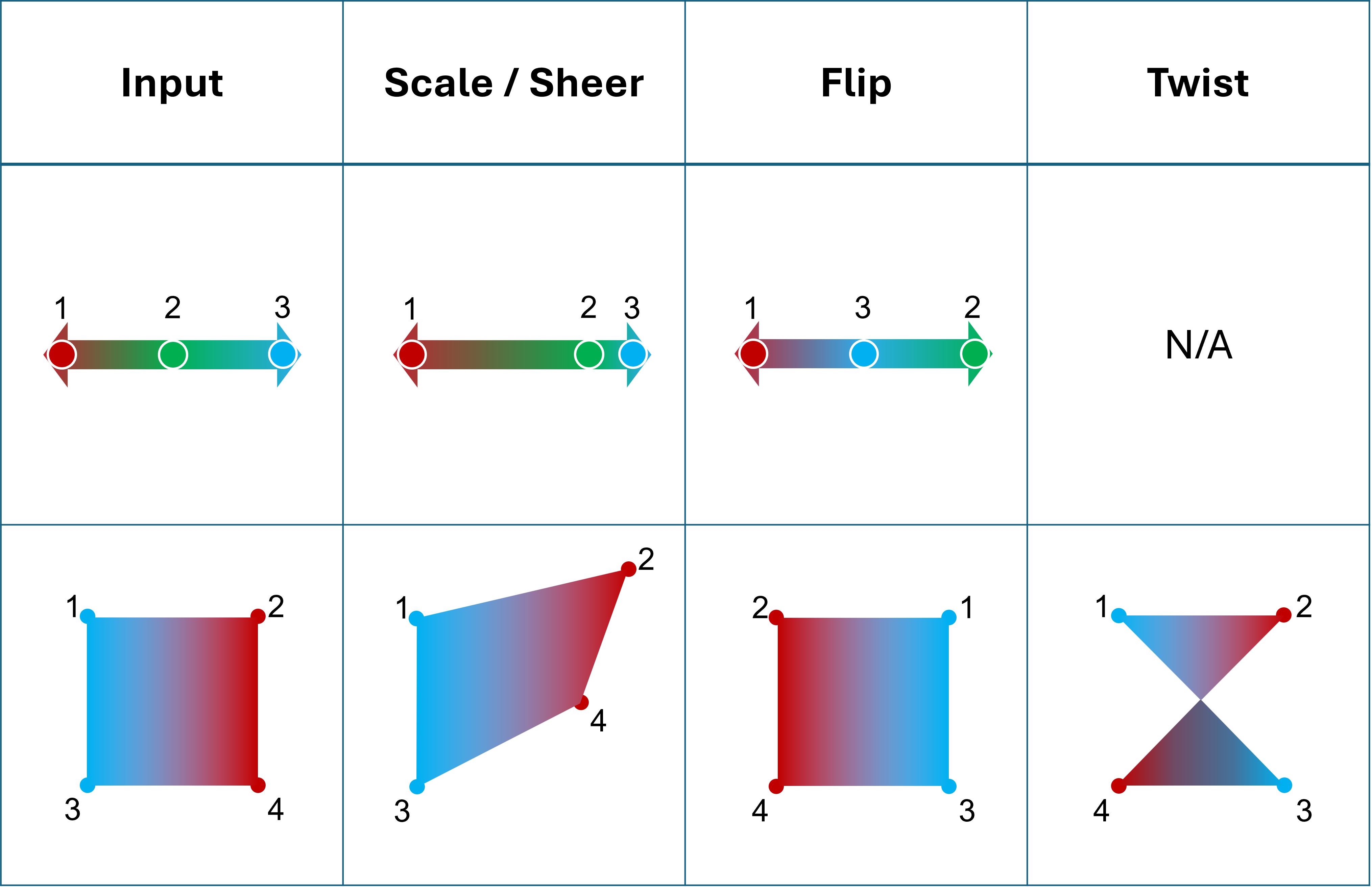}
        \subcaption{Domain operations on 1D/2D inputs.}
    \end{subfigure}\hfill
    \begin{subfigure}{.475\linewidth}
        \includegraphics[width=\linewidth]{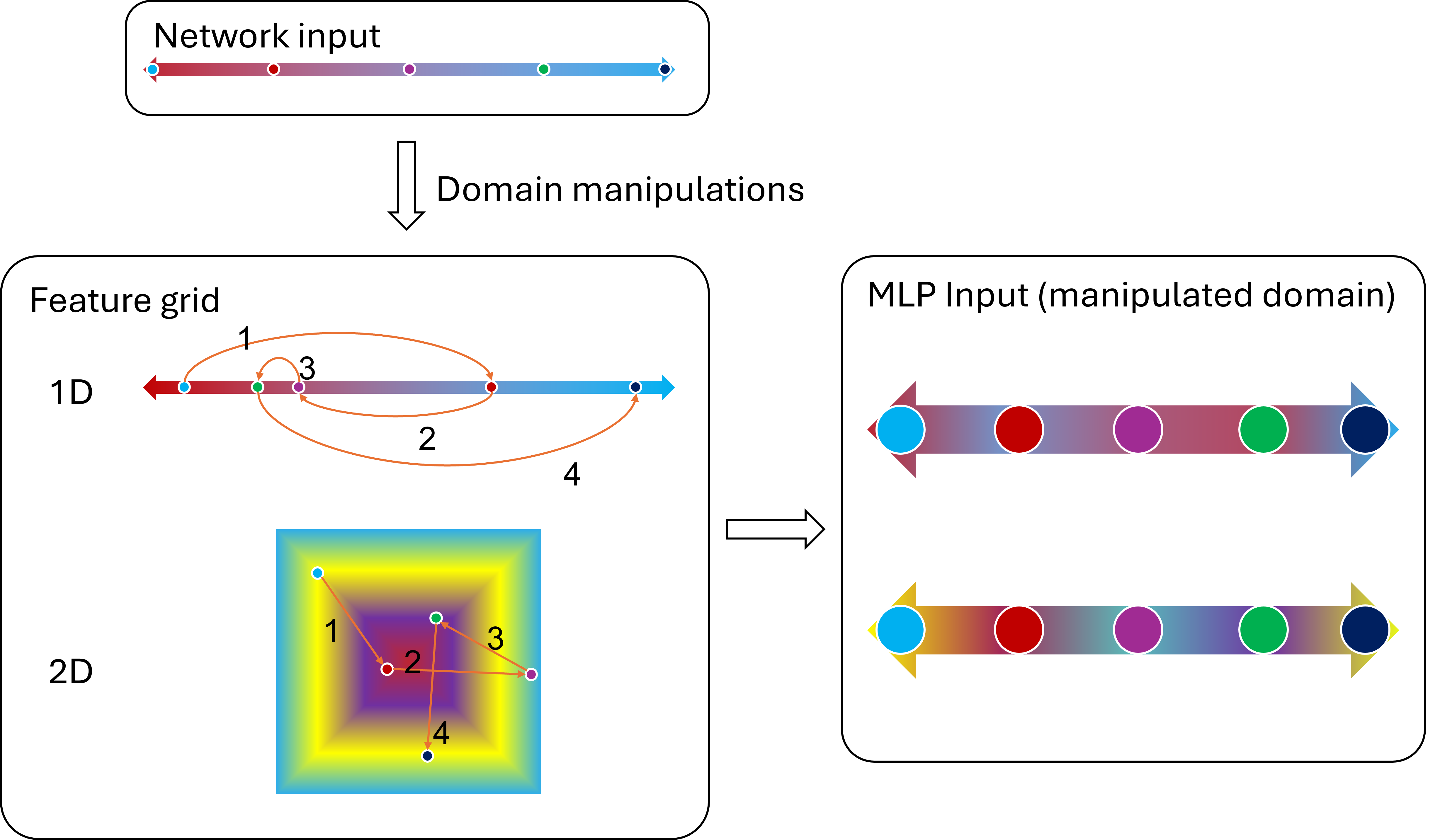}
        \subcaption{Effects of domain manipulation on network inputs.}
    \end{subfigure}
    \caption{Illustration of domain manipulation at different input and feature dimensions.}
    \label{fig:domain_manip_illust}
\end{figure*}

\begin{figure}[h!]
    \centering
    \includegraphics[width=\linewidth]{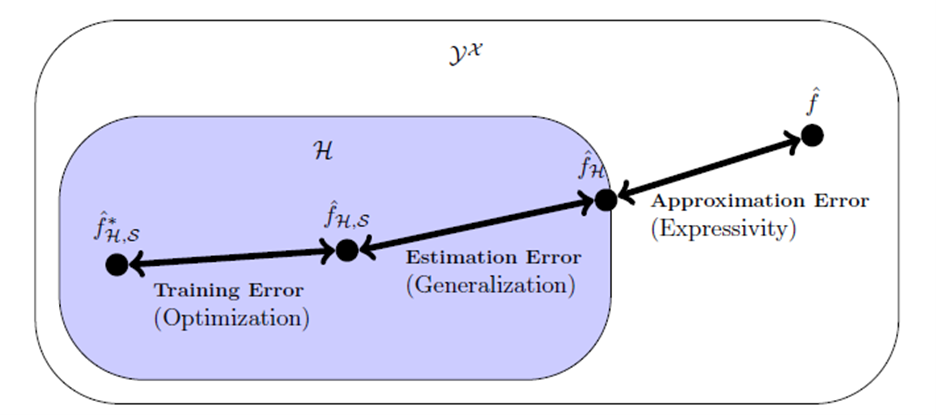}
    \caption{Components of the error between the learnt function and the target function. Taken from \cite{raghu2017expressive}.}
    \label{fig:error_illust}
\end{figure}

\section{Related Works and Background}
\textbf{Grid-based Neural Fields} Before grid-based neural fields, the state-of-the-art was dominated by purely-implicit methods. Most methods either improved the network architecture by introducing better inductive biases such as \cite{2020siren}\cite{lindell2022bacon}\cite{shekarforoush2022residual}\cite{tancik2020fourier}\cite{saragadam2023wire} or divided the problem into numerous smaller subtasks for multiple networks \cite{reiser2021kilonerf} \cite{hao2022loe} \cite{saragadam2022miner}. Plenoxels \cite{fridovich2022plenoxels} was among the first to propose the use of an explicit grid structure to store position-dependent features in the context of NeRFs. It was then followed by numerous efforts to combine explicit grids with implicit neural networks to combine the benefits of both worlds into a hybrid representation such as NGLOD \cite{takikawa2021neural} which proposed the use of a multi-resolution grid, NGP \cite{muller2022instant} which proposed the use of hash encodings, and Neuralangelo \cite{li2023neuralangelo} which proposed the use of numerical gradients across the grid boundaries. To further improve the scaling capabilities of grid-based neural fields, \cite{cao2023hexplane}\cite{chen2022tensorf}\cite{Fridovich-Keil_2023_CVPR} replaced the dense multi-resolution grid with factorized planes.

\textbf{Error Of A Neural Network} \cite{raghu2017expressive} decomposed the total error of a learn function into three components, namely (1) the \textit{approximation error (expressivity)} which is the lower bound error of the best possible approximation of a given network configuration; (2) the \textit{estimation error} which is the difference between the network's theoretically best-approximating function and its data-dependent function; and finally (3) the \textit{training error} which is the difference between the best empirical network and the network that is learned via non-convex optimization in practice. We follow this taxonomy and, similar to a lot of prior work, limit our theoretical exploration to understanding how a grid could improve the \textit{approximation error (expressivity)} of a neural field. This allows us to avoid considering the complex and uncertain effects of how a neural network is being optimized. 

\textbf{Expressivity} One of the earliest expressivity results dates back to the 1989 paper which proposed the Universal Function Approximation Theorem of multilayer perceptions \cite{hornik1989multilayer}. The following works extended the analysis to a wider variety of neural networks with different activations such as Sigmoid \cite{barron1993universal}, ReLU \cite{yarotsky2017error}, and Sin \cite{yuce2022structured}. Expressivity is an arbitrary term that specifies the ``amount'' functions that a network can represent. Mathematically it is generally referred to the spanning set of functions of a given network configuration. Just like how a subspace of the function space could be defined with infinitely many unique basis functions, there are numerous ways to define expressivity in practice. For instance, \cite{yarotsky2017error} related expressivity to the maximum number of linear pieces that a ReLU network can encode, while \cite{yuce2022structured} measures expressivity with the Fourier support of a SIREN network. Given that grid-based neural fields are usually configured with a ReLU MLP, we found it best to define expressivity in terms of the maximum number of linear pieces. As we will show in SECTION \ref{sec:grid}, the effects of the grid also translate directly to an increase in the maximum of linear segments.

\begin{figure*}[htp!]
   \centering
    \begin{subfigure}{\linewidth}
        \includegraphics[width=\linewidth]{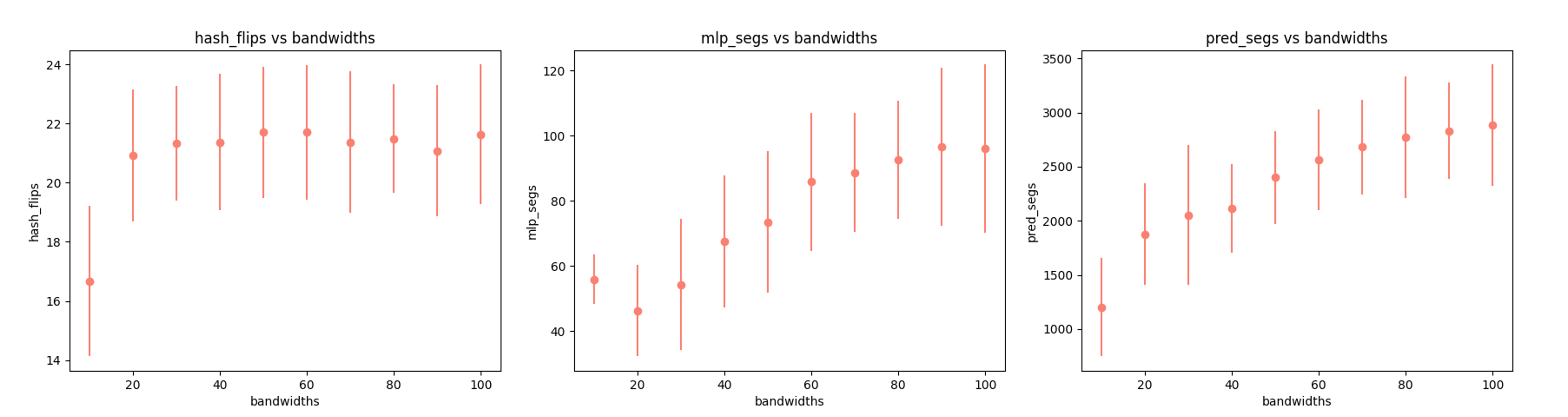}
        \subcaption{NGP.}
    \end{subfigure}
    \begin{subfigure}{\linewidth}
        \includegraphics[width=\linewidth]{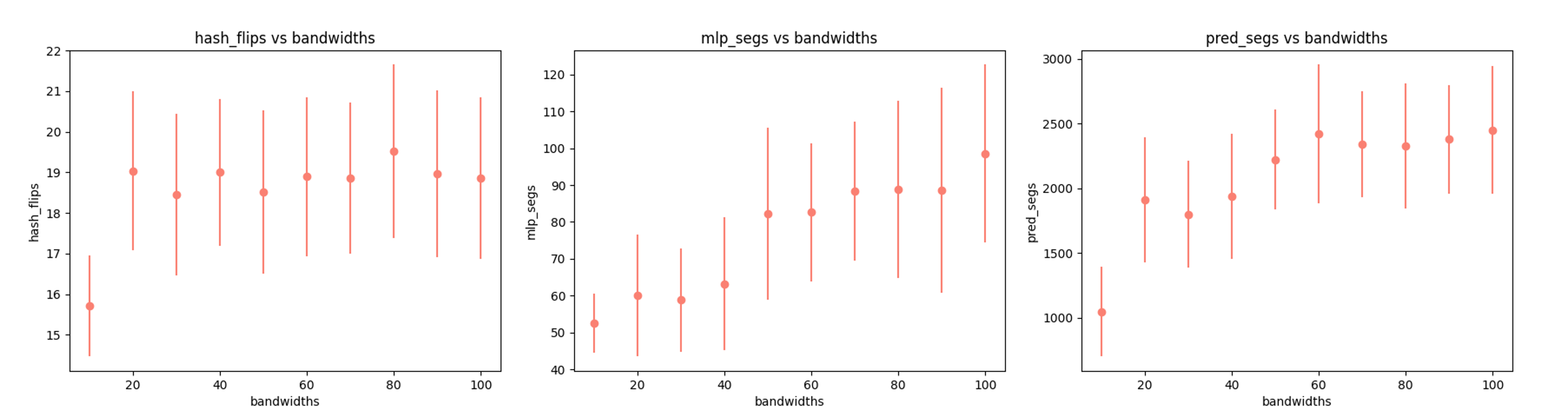}
        \subcaption{NGP with ordered hash initialization.}
    \end{subfigure}
    \caption{Number of domain flips, MLP segments, and prediction segments when fitting to different signal frequency bandwidths.}
    \label{fig:bandwidth_and_flipping}
\end{figure*}

\section{Method - measuring the number of linear segments as expressivty}

In this section, we briefly justify the use of ``the number of linear segments'' as a measurement of expressivity in our analysis. \cite{yarotsky2017error} breaks it down nicely as follows:
\begin{enumerate}
    \item Virtually all ``nice'' function has a Taylor approximation accurate to a sufficient error bound. Note that the Taylor approximation is an infinite series of monomials/polynomials.
    \item It can be shown that the function $f(x) = x^2$ can be approximated accurately by a sufficiently large ReLU network.
    \item It can also be shown that there exist ReLU networks that can approximate the product of two variables, i.e. $g(x, y) = xy$.
    \item Combine (2) and (3), we get that there is always a wide/deep enough ReLU network that can approximate a function with a Taylor approximation to a given error bound.
\end{enumerate}

Note that ReLU networks can only represent piecewise linear functions. In other words, all the requirements stated above about the size of the ReLU network revolve around the maximum number of piecewise linear segments that it can encode. Given the maximum number of piecewise linear segments, we can use error formulas for linear splines to estimate the approximation error of the ReLU network.

\section{Deciphering the effects of the grid}
\label{sec:grid}

\subsection{Working principle of the grid}
\begin{figure}[h]
    \centering
    \includegraphics[width=\linewidth]{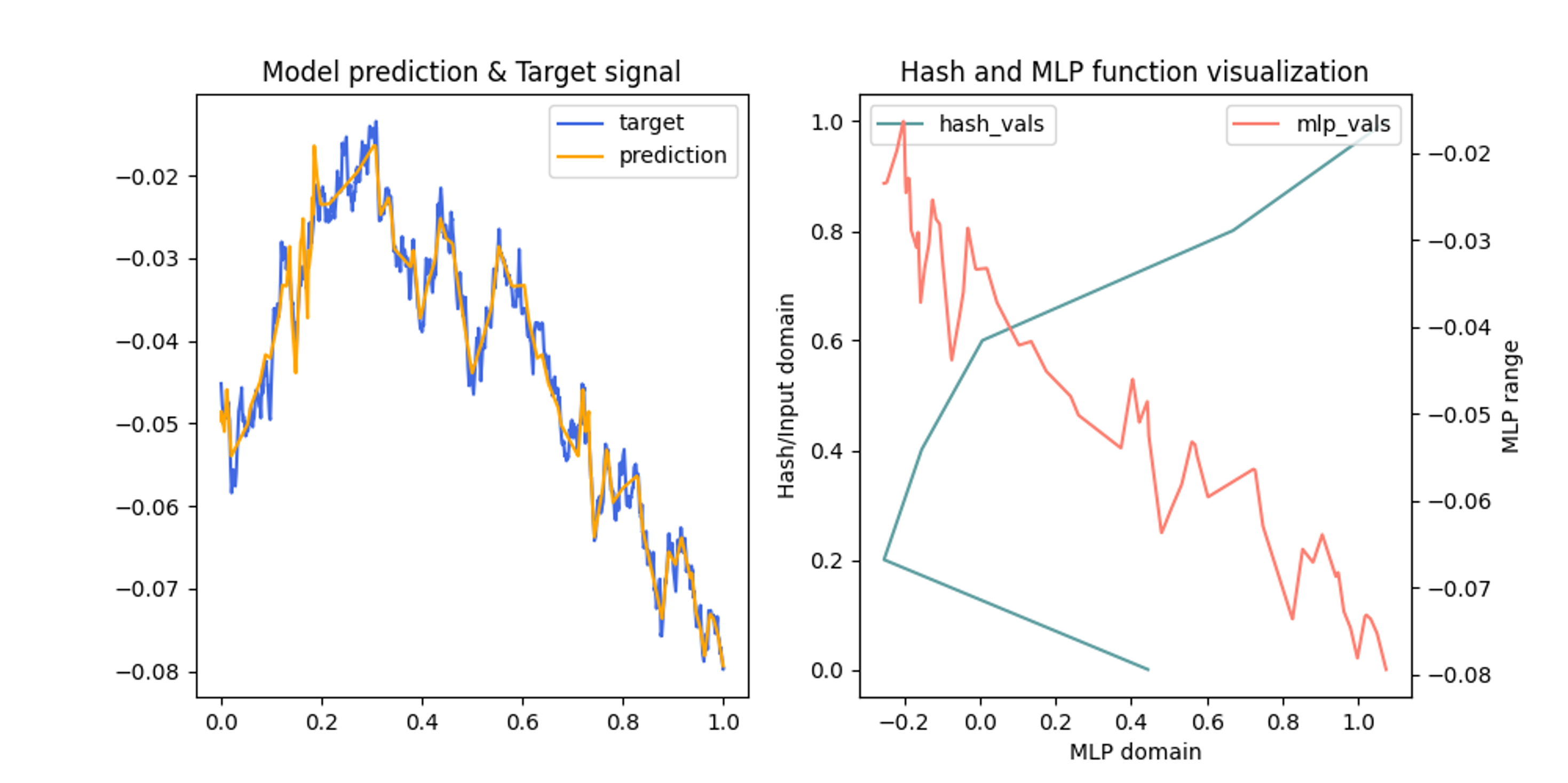}
    \caption{Visualizing the effects of a 1-dimensional, single-resolution, no-hashing grid when fitting to a 1-dimensional signal.}
    \label{fig:grid_1d}
\end{figure}

We build our analysis of NGP incrementally and start with the bare-bones effects of a 1-dimensional, single-resolution, no-hashing feature grid on a 1-dimensional signal. The grid partitions the 1-dimensional input domain into equal-lengthed segments, where the endpoints of each segment (i.e. vertices of the grid) get remapped to a new value in the MLP's domain. The interior of the segments is interpolated linearly between the values of the endpoints. \textbf{The mapping of the grid could hence be understood as two orthogonal operations on the input domain: (1) scaling up/down; (2) ``flipping''.} Take the visualization in Fig \ref{fig:grid_1d} as an example, where the domain $[0,1]$ is partitioned into 5 equal parts. The red line in the right figure is the function that the MLP is representing, while the blue line is a visualization of the grid values, with the $x$-axis being the MLP domain and the left $y$-axis being the input domain. The first segment $[0,0.2]$ is an example of \textbf{flipping}, where the domain gets remapped to start at $\sim0.45$ and ends at $\sim-0.20$. On the other hand, the second and third segments are examples of \textbf{scaling down} where the segment $[0.2, 0.6]$ gets mapped to a very narrow domain between $[-0.2, 0.0]$.

\begin{figure}[h!]
    \centering
    \includegraphics[width=.75\linewidth]{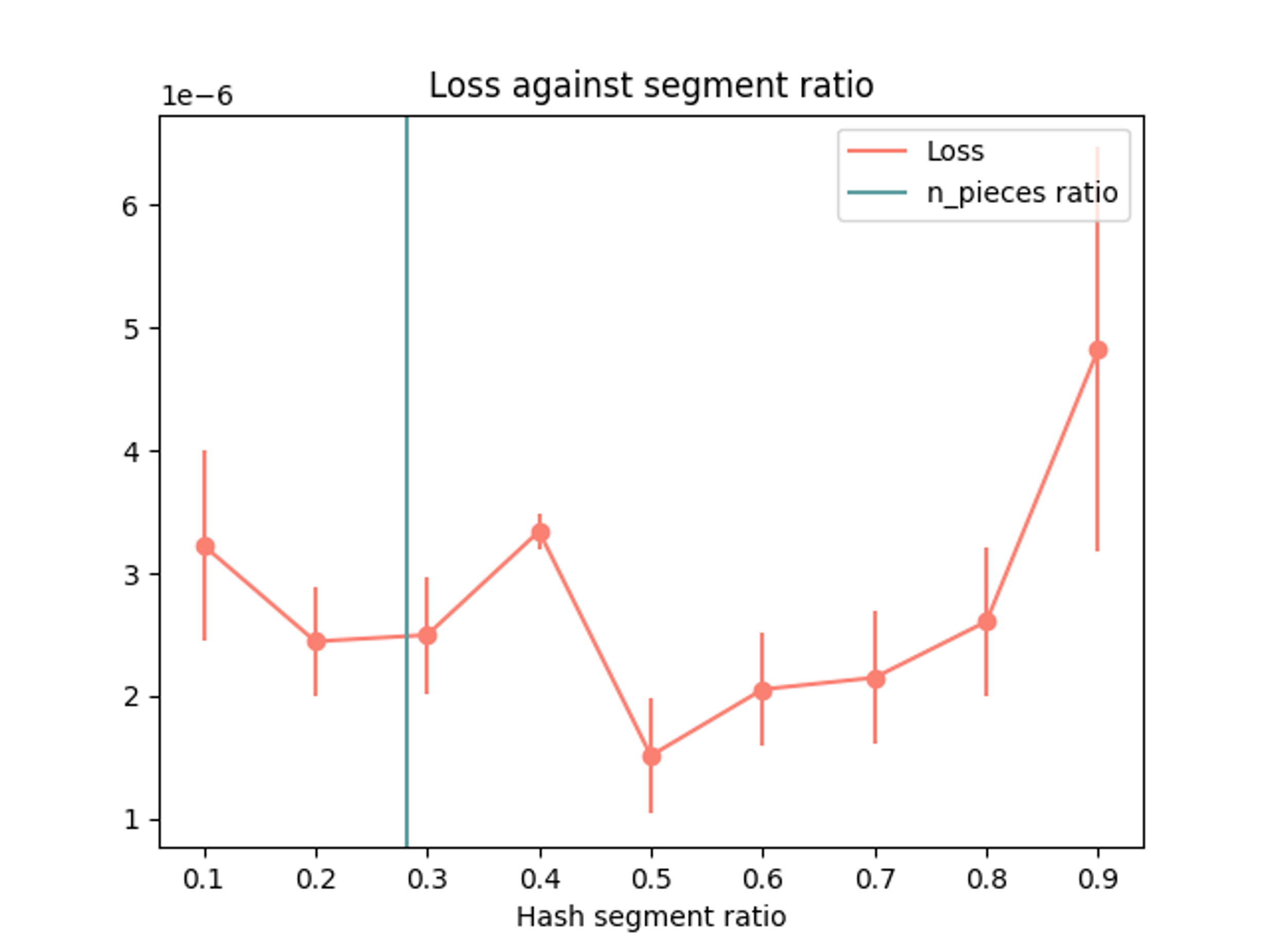}
    \caption{The effect of scaling a segment on the approximation error.}
    \label{fig:scale_up}
\end{figure}

\textbf{Scaling up/down} on itself has very little effect on the network's expressivity. While it is understood that scaling up a relatively complex segment could help distribute more expressive power from the MLP, our experiments show that the model's approximation error is agnostic to the distribution of linear segments along the domain. We fitted a single-layer no-hashing NGP with a resolution of 2 (i.e. splits the domain into two with three feature values) on a target signal that has two segments split at the middle $x=0.5$, each with a significantly different total number of linear segments (i.e $\sim 1:3$ ratio). We fix the first and last hash value to $0.0$ and $1.0$ respectively and increment the center hash value from $0.1$ to $0.9$. We then train the MLP on top of this preset hash table. Fig. \ref{fig:scale_up} summarizes the respective losses of the converged networks. The blue line is the ratio of the number of linear segments in the left portion vs the right portion. Should it be true that the more complex the signal the greater the scale-up of the domain there is, then the optimal hash grid segment ratio should also be near $\sim 0.3$, but Fig. \ref{fig:scale_up} reflects otherwise. It could be understood that the model's expressivity is solely determined by the total number of linear segments that it could encode, but not related to where the segments occur and how densely they are packed.

\begin{figure}[h!]
    \centering
    \begin{subfigure}{.3\linewidth}
        \includegraphics[width=\linewidth]{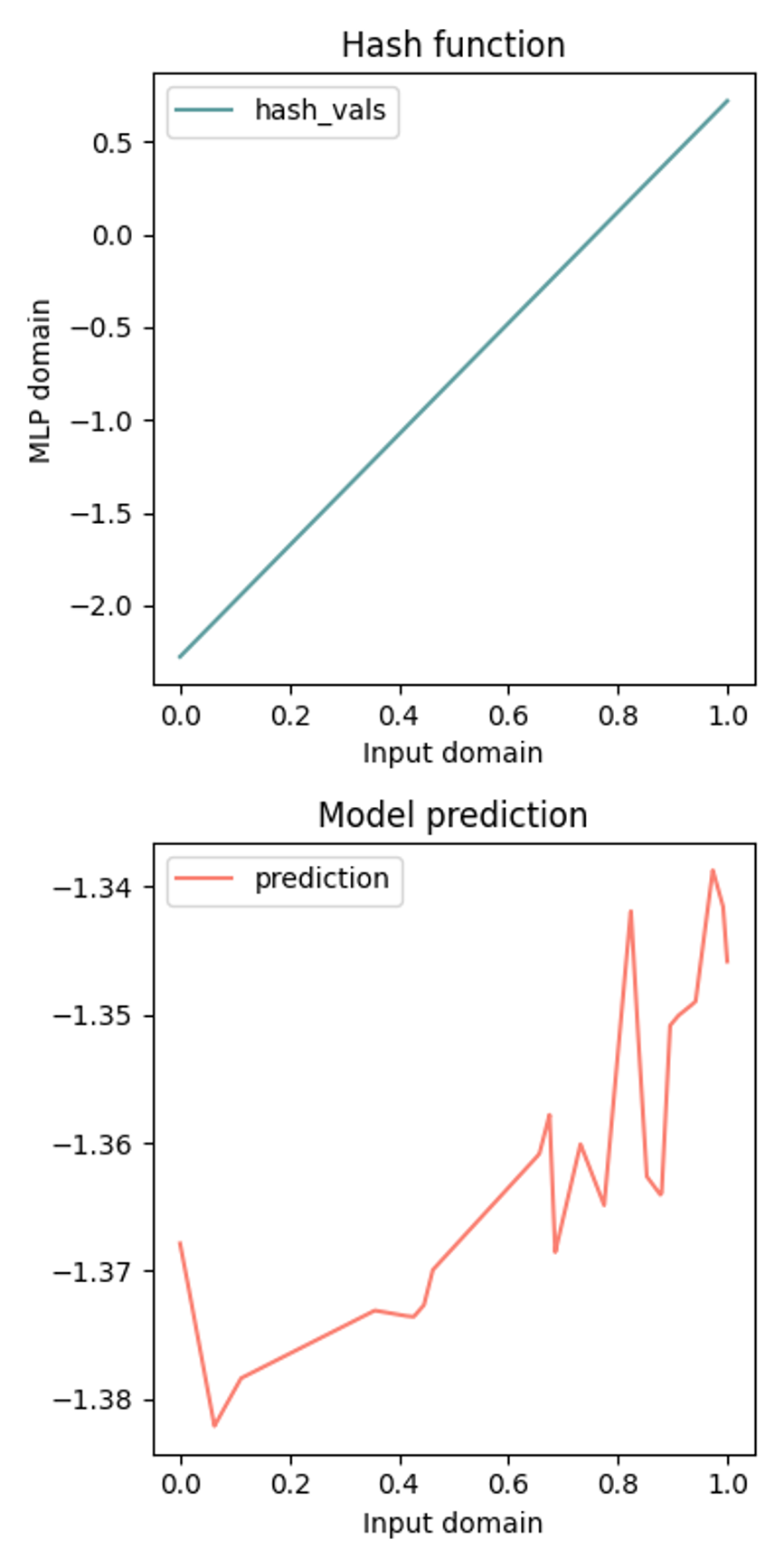}
        \subcaption{No grid.}
    \end{subfigure}
    \begin{subfigure}{.3\linewidth}
        \includegraphics[width=\linewidth]{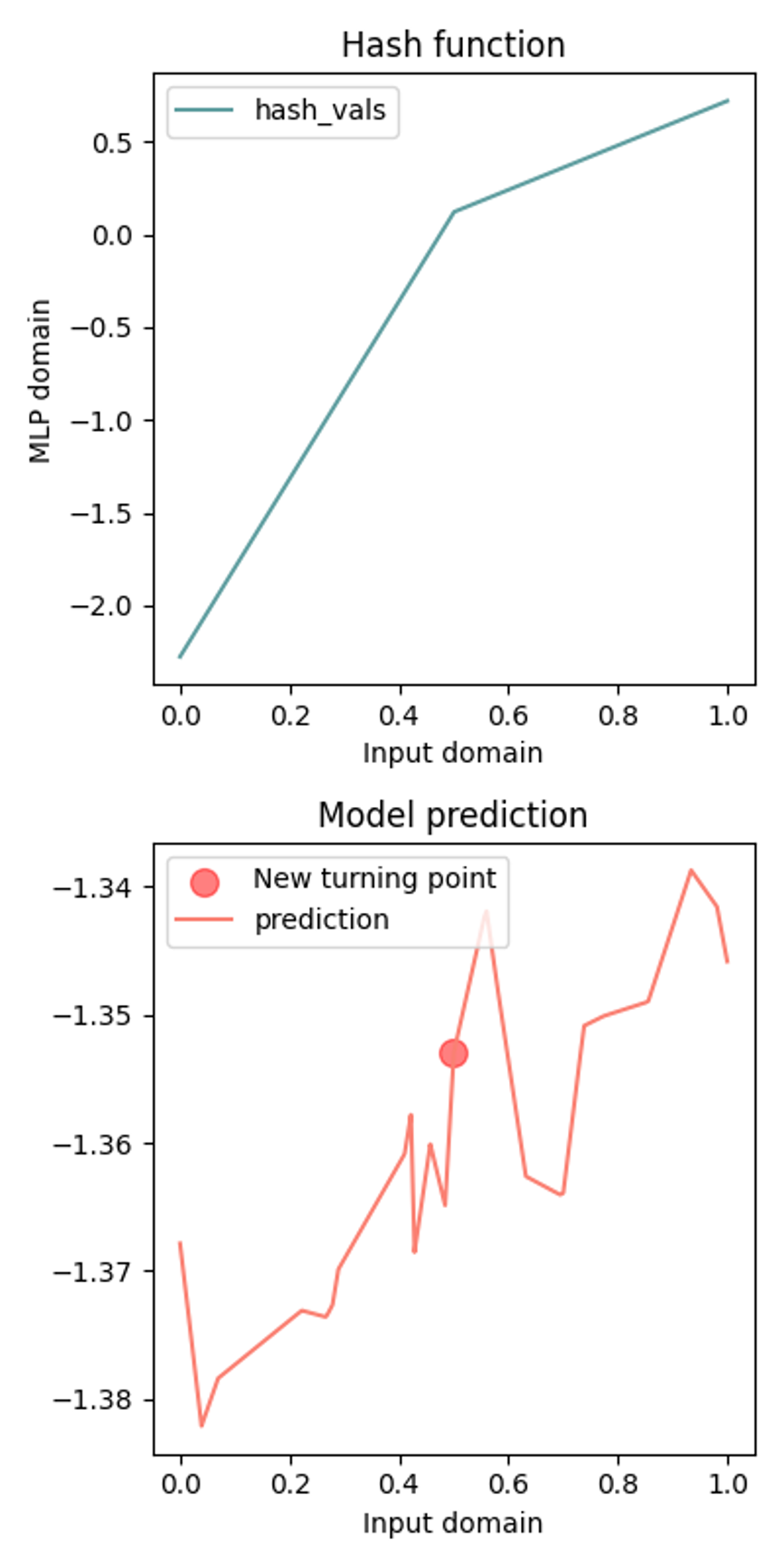}
        \subcaption{Scale.}
        \end{subfigure}
    \begin{subfigure}{.3\linewidth}
        \includegraphics[width=\linewidth]{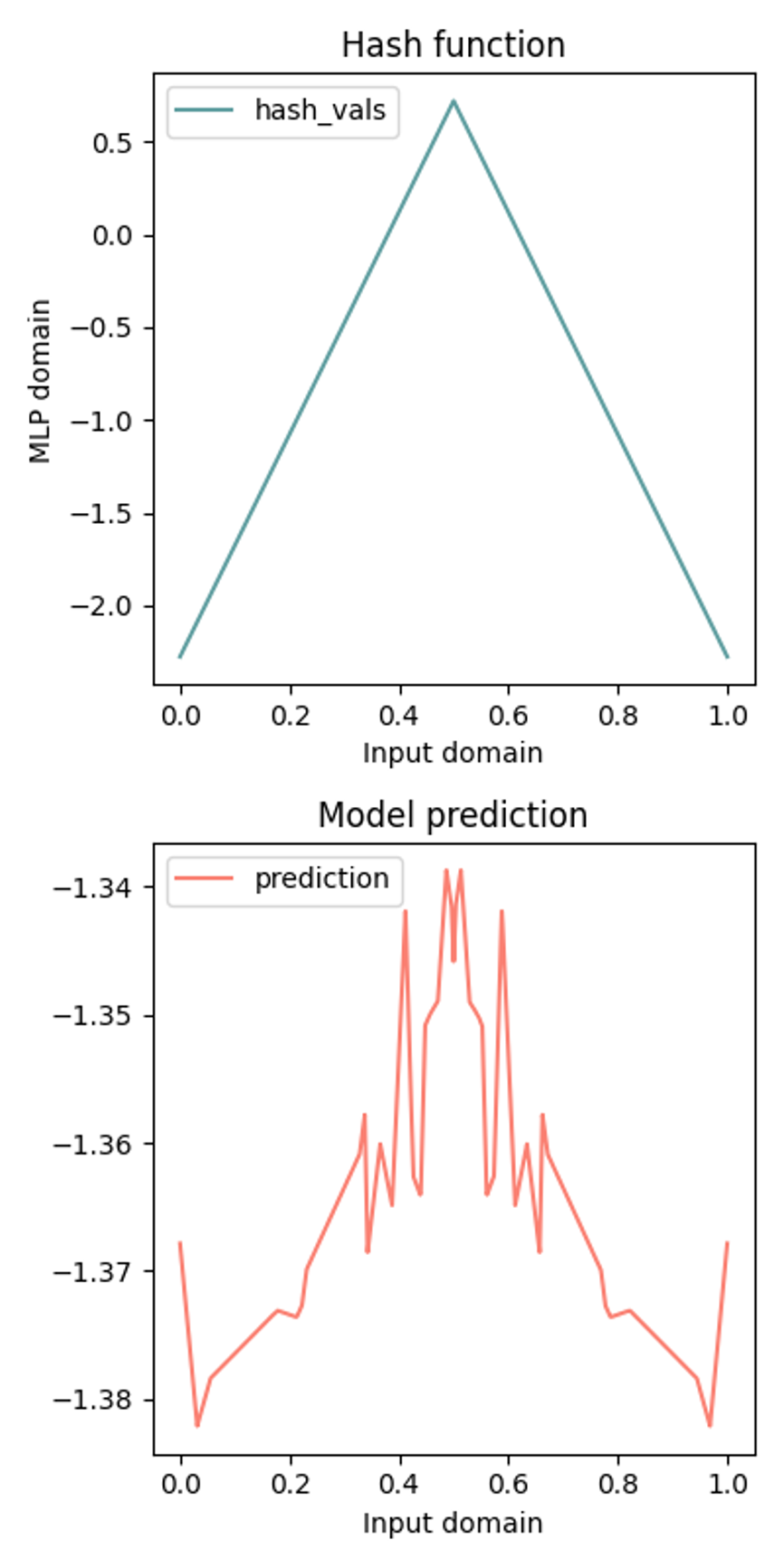}
        \subcaption{Flip.}
    \end{subfigure}
    \caption{Visualizing how the grid's domain manipulation adds new turning points to the network. Top: feature values in a grid with 3 vertices. Bottom: resulting network function.}
    \label{fig:scale_flip_turning_points}
\end{figure}

\textbf{Flipping} however, is a significantly more important operation. Note that the MLP used in NGP only has 2 hidden layers and 64 hidden units. Compared to the signal complexity that NGPs are usually put against (e.g. large NeRF scenes), the grid must be doing a lot more than just redistributing expressivity across segments as the MLP simply does not have sufficient expressivity as a whole. As mentioned previously, an MLP's expressivity is directly determined by the maximum number of linear segments that it can encode. We now try to translate the two grid operations into changes in this upper bound. Fig. \ref{fig:scale_flip_turning_points} presents a toy visualization of this. Firstly, at each grid vertex, if the two neighboring segments are scaled to different ranges, then an additional turning point is added to the composite network, on top of what is already provided by the MLP. Hence, scaling up/down permits an addition of $N_{res}$ linear segments to the network, where $N_{res}$ is the resolution of the grid. On the other hand, when neighboring segments are flipped, entire segments of the MLP which may contain multiple linear segments could be reused. This implies that flipping permits a maximum of $N_{res} \times N_{mlp}$ additional linear segments to the network, where $N_{mlp}$ is the number of linear segments encoded in the MLP. It is also important to note that the importance of the scale up/down operations only becomes apparent when it is accompanied by the flipping operations. While in itself it could not add turning points to the network effectively, it is an essential tool for tuning how much of the two neighboring segments are to be overlapped in the MLP domain. If only 10\% of the second segment mirrors the entirety of the first segment, then during the flipping, the second segment should also be scaled by 10$\times$.

\begin{figure}[h!]
    \centering
    \includegraphics[width=\linewidth]{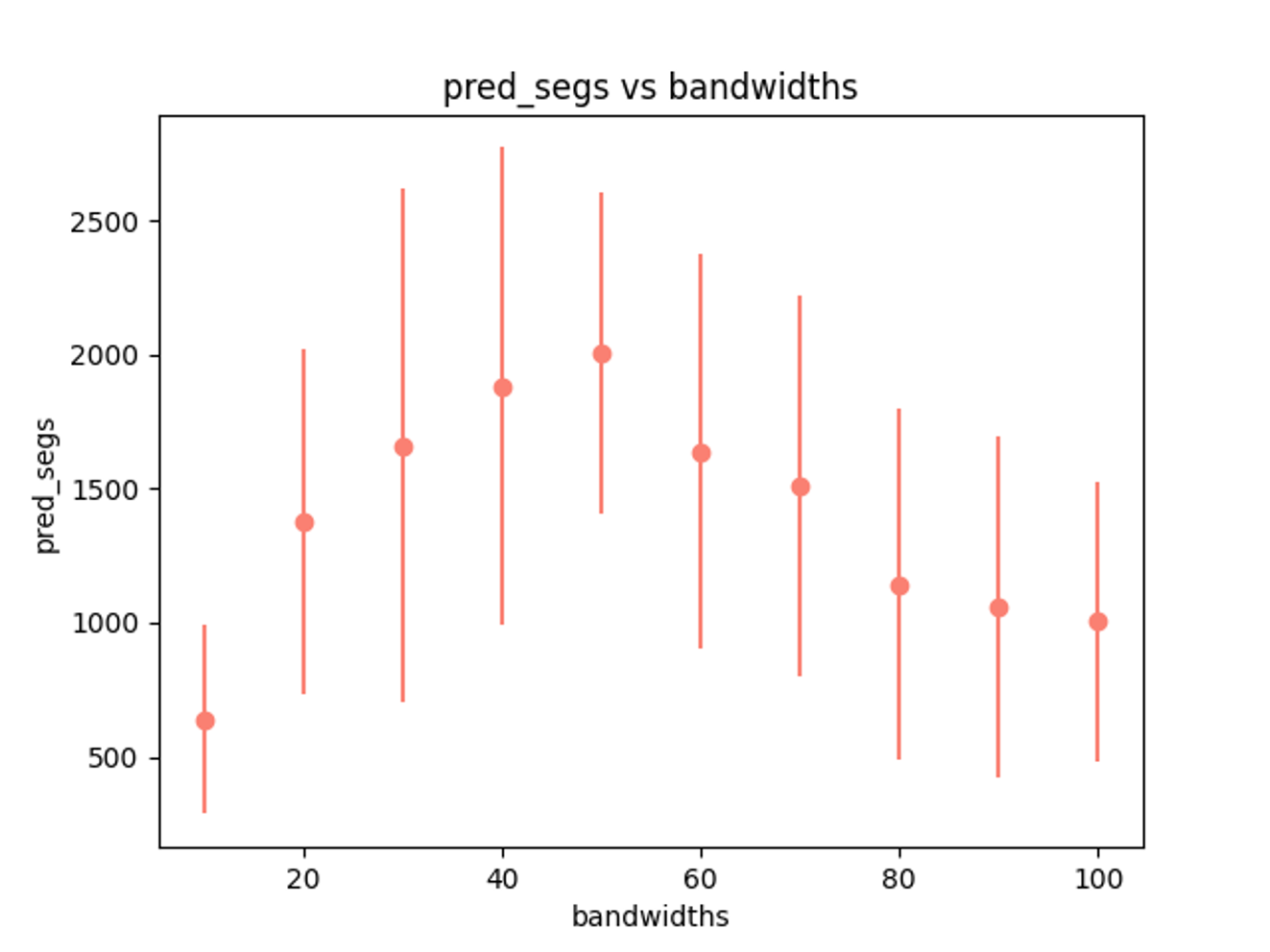}
    \caption{Number linear segments in a ReLU network when fitting to different signal frequency bandwidths.}
    \label{fig:relu_segs}
\end{figure}

The superior effectiveness of increasing expressivity by domain flipping is also reflected in practice. Fig. \ref{fig:bandwidth_and_flipping} shows a summary of the number of domain flips, MLP linear segments, and prediction linear segments when fitting to signals of different frequency bandwidths. In particular, we generated 3 different signals with randomly sampled coefficients of the first 100 frequencies and zeroed out the coefficients from high to low with increments of 10 frequencies. We trained two different sets of NGPs, one with randomly initialized weights for both the grid and the MLP, and another where we initialized the grid with ordered values. All models have a single-level grid of resolution 25 and an MLP of 4 hidden layers and 64 hidden units. The leftmost column of Fig. \ref{fig:bandwidth_and_flipping} shows how the grid almost always converges to having more flips than scaling (19-21 out of 25 grid vertices are turning points). This is the same case even when the grid is initialized to be ordered, as shown in the second row. The only exception is when the signal bandwidth is 10, which we believe is because the number of flips in the signal itself is less than 25. Fig. \ref{fig:bandwidth_and_flipping} also shows empirically how the grid has a multiplier effect on the number of linear segments in the prediction signal as we compare the $y$-axis of the second and third columns. This results in a total number of linear segments from $\sim$2k to 2.5k, increasingly with the target frequency bandwidth, which is generally higher than that of a vanilla MLP of identical configurations as shown in Fig. \ref{fig:relu_segs}.

Despite domain flipping's effectiveness, it is worth noting its constraints and trade-offs. Note that whenever domain flipping occurs, segments of the domain may be reused as input to the MLP to fit different parts of the target signal. This implies that repeated segments of the prediction signal must exhibit similar shapes to every one of its corresponding target signals or their mirrored versions. Otherwise, approximation error would incur. In other words, this would require the target signal to have neighboring segments that are approximately mirrored. However, for complex signals in the wild, this form of self-similarity rarely occurs in a convenient form for the grid to exploit.


Considering how often the grid flips the domain as shown in Fig. \ref{fig:bandwidth_and_flipping}, we hypothesize that the grid converging to a flipping configuration may be a result of easier convergence, but not necessarily a globally optimal solution. That is, since domain flipping artificially creates multiples of turning points for the prediction signal, it serves as an ``easier'' method to fit the target signal compared to optimizing the weight matrices of the MLP. Indeed, there are instances where we could see that the occurrences of the grid flipping early in the training stage lead to the model getting stuck at local minima. However, since this analysis involves more considerations about the network's convergence characteristics, we leave this as an open-ended direction for future works.


\subsection{The grid at higher dimensions}
The idea of domain manipulation extends naturally to higher dimensions and unlocks more flexibility in manipulating the MLP domain as each piecewise linear segment has exponentially more boundary vertices. For instance, when $N = F = 2$, the original scale-up/down operations are converted into a more expressive version of \textbf{sheering}, and the flipping operation splits into \textbf{surface flipping} and \textbf{surface twisting}. Some examples of this are visualized in Fig \ref{fig:domain_manip_illust} (a). These additional operations create more options to repeatedly utilize the linear segments of the MLP, deeming the grid a more effective method to increase the general expressivity of the overall network. the addition of operations such as twisting also reduces the ``rate'' and ``area'' at which segments would have to overlap, which reduces the negative effects of the two aforementioned restrictions.

However, it is only when we increase the feature dimension $F$ to greater than $N$ that these constraints are completely alleviated as it removes the necessity of having overlapping MLP segments to save expressivity. As shown in Fig \ref{fig:domain_manip_illust}(b), a grid with 1-dimensional input traces a piecewise linear path on a 2-dimensional manifold. When $F > 1$, each segment will rarely intersect with another segment entirely but rather would intersect at most at a single point. In fact, for any $N$-dimensional input, as long as $F > N$, segment overlaps would rarely occur.

In practice, the authors of NGP have found it to be a good balance between performance and parameter count when the feature dimension $F$ is fixed at 2, regardless of the input dimension. While our analysis shows that such a grid may be deemed far less effective if we are working with 2D or even 3D data, we highlight that having a multi-resolution grid structure has a similar effect as having higher dimensional features. This similarly solves the problem of segment overlaps by having an MLP input dimension of $L \times F = 4$, which is larger than the network input dimension.

\section{Conclusion and future works}
In this paper, we proposed a novel perspective, namely \textbf{domain manipulation}, to understand the effectiveness of Instant-NGP, which is a state-of-the-art neural field that is characterized by its multi-resolution hash grid. This perspective provided a ground-up explanation of how the use of a feature grid increases the expressivity of the neural field, leading to superior reconstruction performances. Empirical visualizations from 1-dimensional target signals aided the development of our perspective. We focused our analysis on expressivity only to avoid numerous uncontrollable variables when considering network convergence (e.g. choice of optimizer, learning rates, schedulers, etc.).

As we mentioned in the introduction, Instant-NGP's effective architecture is made up of both the feature grid and a multi-resolution hashing solution. While our analysis provided an intuitive understanding of the feature grid, we do note that the relation between the hashing solution and the neural field's expressivity is yet to be explored. It is understood that the combination of a multi-resolution grid and the MLP serves as an effective hash collision resolution. However, it is unclear how the MLP could serve both as the underlying signal provider and a hash collision resolver. It is also unproven that the multi-resolution grid structure is the optimal architecture of a parameter-to-expressivity trade-off. Lastly, we highlight that one of the greatest attributes of Instant-NGP is its fast convergence (even without fully-fused CUDA pipelines). We believe that the multiplier effect of the grid on the MLP's number of linear segments could be a reason for the ease of convergence, as it is generally harder to optimize the fully-connected weights of the MLP than to optimize the independent feature values in the grid. A study of the convergence properties of Instant-NGP is also of great importance for future works.

\bibliographystyle{IEEEtran}
\bibliography{ref}






\end{document}